\documentclass[letterpaper, 10 pt, conference]{ieeeconf}  

\IEEEoverridecommandlockouts                              
\usepackage{graphicx} 
\usepackage[cmex10]{amsmath} 
\usepackage{amssymb}  
\usepackage{cite}
\usepackage{bm}
\usepackage{hyperref}
\usepackage{bookmark}
\usepackage{tikz}
\usepackage{lipsum,adjustbox}
\usetikzlibrary{arrows,positioning,decorations.pathreplacing,shapes}

\newtheorem{definition}{Definition}

\newtheorem{proposition}{Proposition}

\DeclareMathOperator*{\argmin}{arg\,min}

\newcommand{\Uff}{U_{\mathrm{ff}}}

\newcommand{\uff}{u_{\mathrm{ff}}}

\title{\LARGE \bf
Model Predictive Trajectory Optimization and Tracking for On-Road Autonomous Vehicles
}

\author{Peng Liu, Brian Paden, and Umit Ozguner
\thanks{P. Liu and B. Paden are with the Advanced Technology Group,
        Samsung, San Jose, CA 95134, USA
        {\tt\small peng2.liu@samsung.com, brian.paden@samsung.com}}%
\thanks{U. Ozguner is with the Department of Electrical and Computer Engineering, The Ohio State University, Columbus, OH 43210, USA
{\tt\small ozguner.1@osu.edu}}%
}

\begin{document}

\maketitle
\thispagestyle{empty}
\pagestyle{empty}

\begin{abstract}
Motion planning for autonomous vehicles requires spatio-temporal motion plans (i.e. state trajectories) to account for dynamic obstacles.
This requires a trajectory tracking control process which faithfully tracks planned trajectories.  
%
In this paper, a control scheme is presented which first optimizes a planned trajectory and then tracks the optimized trajectory using a feedback-feedforward controller.
The feedforward element is calculated in a model predictive manner with a cost function focusing on driving performance. Stability of the error dynamic is then guaranteed by the design of the feedback-feedforward controller.
%
The tracking performance of the control system is tested in a realistic simulated scenario where the control system must track an evasive lateral maneuver.
The proposed controller performs well in simulation and can be easily adapted to different dynamic vehicle models. 
The uniqueness of the solution to the control synthesis eliminates any nondeterminism that could arise with switching between numerical solvers for the underlying mathematical program.


\end{abstract}

\section{INTRODUCTION}

Autonomous vehicles often decompose the selection of steering, throttle, and braking control signals into a planning process which generates a feasible motion through the perceived scene. This is followed by a control process which executes a local trajectory tracking control policy robust to process noise and load disturbances.
%
Motion planning algorithms for autonomous driving usually simplify the planning task by first planning a geometric path followed by planning a longitudinal velocity profile along the geometric path~\cite{xu2012real,ziegler2014making}.
%
%
Since all motion planning algorithms with approximate completeness guarantees have exponential complexity with respect to state dimension, this decomposition affords significant reduction in computational requirements and planning latency.
The decomposition is inherited by the control system which generally has separate lateral and longitudinal control policies~\cite{paden2016survey}. 
%
%
The subtleties of designing controllers in this case have been well studied~\cite{rupp2017survey,snider2009automatic}.
%
%
%
%

The downside to this decomposition is that it implicitly discards effective options available to the planner.
For example, if a dynamic obstacle suddenly crosses the vehicle's path, a geometric planner will not reason about the motion of that object and will have to assume a fixed location, or neglect the dynamic obstacle completely; the subsequent longitudinal planner must find a safe option restricted to the selected geometric path.
%
%
This motivates spatio-temporal motion planning that accounts for predicted states of dynamic obstacles.
However, the prohibitive complexity of motion planning for high fidelity models~\cite{canny1988complexity} forces motion planning modules to use simplified dynamic models with low dimensional continuous state spaces, sometimes further approximated by finite state models~\cite{paden2016survey,lavalle2006planning}. 
This places a greater burden on the control process which must execute a trajectory tracking control policy that not only compensates for load disturbances and accounts for sensor noise, but has to account for model difference between the planner and the vehicle's dynamics. 

%
%
%
%
%

%
To address such issue, one approach attracting increasing attention is Model Predictive Control (MPC) which incorporates a sophisticated dynamic model to recursively optimize tracking errors over the reference trajectory.
%
%
%
%
%
%
%
%
MPC approaches for trajectory tracking have been widely investigated for both autonomous driving \cite{ziegler2014making,raffo2009predictive} and semi-autonomous driving cases \cite{gray2012predictive,falcone2007predictive}. 
Furthermore, if the cost function can be proved as a Lyapunov function, stability of the nominal closed-loop system is guaranteed~\cite{mayne2000constrained}. 
However, the cost function for trajectory tracking of autonomous vehicles typically need to account for riding performance such as small yaw rate and smooth change of acceleration, which makes proof of the cost function as a Lyapunov function very difficult, even intractable. 

%
%
%

This paper discusses a particular control system design for trajectory tracking which utilizes a feedforward MPC to optimize the reference trajectory with respect to a cost function that does not need to be a Lyapunov function of the closed-loop error dynamic.
%
%
%
%
%
%
The proposed feedforward MPC, discussed in Section \ref{sec:main-result}, is formulated as a strictly convex optimization problem (SCOP), which results in unique solutions. 
%
%
%

\begin{figure*}
\centering
\begin{adjustbox}{width=14cm}
\begin{tikzpicture}[
    squarenode/.style={rectangle, draw=blue!60, fill=green!5, align=center, anchor=west, thick, minimum width=2cm, minimum height=1cm}
] 
\node[squarenode] (predictor) at (-1, 0) {trajectory\\ optimization};
\node[squarenode] (vehicle)  at (3.8,-1.5) {vehicle};
\node[squarenode] (filtering)  at (6.5,-1.5) {filtering};
\node[squarenode] (feedback) at (3.8,-3.0) {feedback \\ controller};
\node[squarenode] (model) at (-1,-2.0) {vehicle\\ model};
\node[squarenode] (planner) at (-5,0) {motion\\ planner};
\draw (4.5,0.49) node[align=center] {nominal trajectory};
\draw (2.3,-0.1) node {feed-forward};
\draw (7.0,-2.8) node[align=center] {state error};
\draw[draw=blue!60, very thick] (2.5,-1.5) circle [radius=0.08cm];
\draw (2.3,-1.7) node {$+$};
\draw (2.7,-1.3) node {$+$};
\draw[draw=blue!60, very thick] (11.0,-1.5) circle [radius=0.08cm];
\draw (10.8,-1.7) node {$+$};
\draw (11.2,-1.3) node {$-$};
\draw[thick,draw=blue!60] (1,0.3) -- (11.0,0.3);
\draw[->,thick,draw=blue!60] (1,-0.3) -| (2.5,-1.45);
\draw[->,thick,blue!60] (2.55,-1.5) -- (vehicle.west);
\draw[->,thick,draw=blue!60] (vehicle.east) -- (filtering.west);
\draw[->,thick,draw=blue!60] (filtering.east) -- (10.95,-1.5);
\draw[->,thick,blue!60] (11.0,0.3) -- (11.0,-1.45);
\draw[->,thick,blue!60] (11.0,-1.55) |- (feedback.east);
\draw[->,thick, blue!60] (planner.east) -- (predictor.west);
\draw[->,thick,blue!60] (feedback.west) -| (2.5,-1.55);
\draw[->,thick,blue!60,dashed] (model.north) -- (predictor.south);
\draw[->,thick,blue!60,dashed] (model.south) |- (3.8,-3.2);
\draw (-2.0,0.45) node[align=center] {reference\\ trajectory};
\draw (9.6,-1.3) node[align=center] {vehicle state};
\draw (3.3,-1.3) node {input};
\end{tikzpicture}
\end{adjustbox}
\caption{Diagram of trajectory optimization and the feedback-feedforward scheme. When a reference trajectory from the motion planner is available, the trajectory optimization generates the feedforward input by solving a strictly convex optimization problem that smooths the reference trajectory. The control input to the vehicle is consist of a feedforward term from the trajectory optimization and a feedback term from the state feedback controller.}
\label{fig:sys-diagram}
\end{figure*}
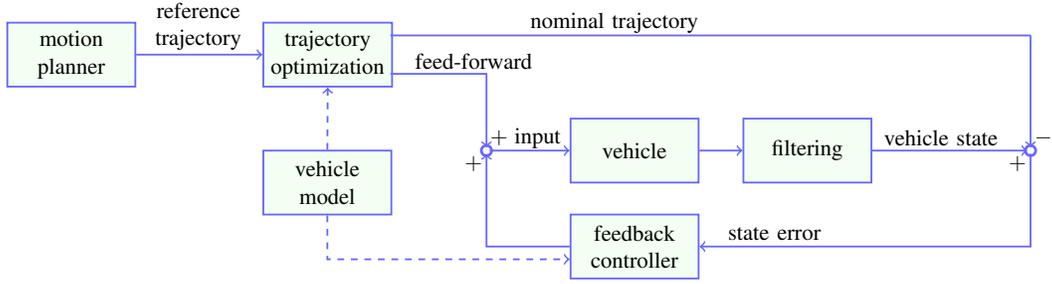

Then, a feedback-feedforward controller is proposed based on a time-varying Linear Quadratic Regulator (TVLQR), where the feedforward element and the nominal trajectory are obtained from the feedforward MPC. 
The reduced computational requirements to run a TVLQR allows the feedback loop to be run at higher frequency than the feedforward path. Moreover, stability of the error dynamic of the nominal system is ensured by carefully choosing coefficients of the cost function in the TVLQR. 

Simulations results in Section \ref{sec:sim} suggest the approach produces satisfactory tracking results in realistic driving scenarios, and with computation times well within the sample rate of the controller.
We conclude in Section \ref{sec:conclude} that this control system design is a promising approach worth further study including in our research teams repertoire of control systems to be tested on or experimental autonomous driving platform.
\section{PROBLEM FORMULATION}
\label{sec:problem}
\subsection{Trajectory Optimization}
\label{subsec:traj-opt}

A reference trajectory, denoted as $\hat{\mathcal{T}}_t:=\{\hat{\mu}_0, \hat{\mu}_{1},\dots, \hat{\mu}_{N}\}$ with  $t\in\mathbb{N}$, published by the planner at instant $t$ is consists of $N+1$ reference states to be reached at uniformly spaced time intervals of duration $\Delta t$.
%
%
The trajectory optimization process is to generate a sequence of control-state pairs  satisfying the dynamic model of the vehicle.
%
We refer to the state sequence and the control sequence as the nominal trajectory and the feedforward term if the trajectory optimization problem is feasible. 
Given the initial state, the vehicle is expected to track the nominal trajectory without error using the feedforward control if no disturbances and model uncertainty appear. 
Fig. \ref{fig:sys-diagram} shows the connections of the trajectory optimization to the motion planner and the feedback controller. 
Instead of the reference trajectory, the optimized nominal trajectory is used by the feedback controller to calculate tracking state-error and to generate the feedback term. 
The feedforward control together with the feedback term formulates the control input that actuates the vehicle. 
The state of the vehicle and feedforward control are denoted $x$ and $\uff$ respectively. 
%
When comparing states in $\hat{\mathcal{T}}$ and $\mathring{\tau}$ is required, we augment $\hat{\mu}$ to $\mathbb{R}^{\mathrm{dim}(x)}$ with the values of the augmented states set to the equilibrium point, and denote the augmented state as $\hat{x}$. Let $\hat{\tau}\triangleq [\hat{x}(t)^T,\dots,\hat{x}(t+N)^T]^T$ be the augmented vector of states from the reference trajectory. The trajectory optimization problem is defined,
\begin{subequations}
\label{eqn:traj-opt}
\begin{align}
   \argmin_{\Uff\in\mathcal U\times\dots\times\mathcal U} & J(\hat\tau, \mathring{\tau}, \Uff)\\
   \text{subject to } & x(t+1) = f_{\mathrm{ff}}(x(t), \uff(t))\\
   & x(t+k) \in \mathcal X, \quad k = 0,\dots,N\\
   & u(t+k) \in \mathcal U, \quad k = 0,\dots,N-1\\
   & \Uff = [\uff(t)^T, \dots, \uff(t+N-1)^T]^T\\
   & \mathring{\tau} = [x(t)^T, \dots, x(t+N)^T]^T\\
   & x(t) = x_0
\end{align}
\end{subequations}
where $f_{\mathrm{ff}}(\cdot,\cdot)$ is the feedforward model of the vehicle, $\Uff$ is the vector of all feedforward terms over the samples from $t$ to $t+N$. $\mathcal{X}$ and $\mathcal{U}$ are the state constraint set and the feedforward input constraint set, respectively.

While tracking performance requirements can imposed with an inequality constraint on error $||\hat\tau - \mathring{\tau}||$, this can make the optimization (1) infeasible. 
Thus, we focus on formulating tracking performance in the cost function as a soft constraint. Detailed construction of the cost function (\ref{eqn:traj-opt}a) and constraints (\ref{eqn:traj-opt}b-d) to a strictly convex quadratic program will be discussed in the following two subsections. 

\subsection{Vehicle Dynamics}


The trajectory optimization problem formulated in \ref{subsec:traj-opt} relies on the equality constraint (\ref{eqn:traj-opt}b) to predict future states and satisfying the differential constraints of the vehicle. 
The trajectory optimization estimates (\ref{eqn:traj-opt}b) from the same dynamic model used in the feedback controller, which is defined as follows
\begin{subequations}
\label{eqn:vedyna}
\begin{align}
\dot s &= v\cos{\theta}\\
\dot y &= v\sin{\theta}\\
\dot \theta &= \frac{v}{L}\tan{\delta}\\
\dot \delta &= -\lambda_1\delta + \lambda_1\delta_{\mathrm{in}}\\
\dot v &= \alpha\\
\dot \alpha &= -\lambda_2\alpha + \lambda_2\alpha_{\mathrm{in}}
\end{align}
\end{subequations}
where $s$, $y$, $\theta$ are the pose at the center point of the rear axle of the vehicle in an inertial  coordinate system. $\delta$, $v$, $\alpha$, $L$ are the steering-wheel angle, the longitudinal speed, the longitudinal acceleration, and the wheel base of the vehicle, respectively. $\delta_{\mathrm{in}}$ and $\alpha_{\mathrm{in}}$ are the control input of steering angle and the control input of acceleration/deceleration, respectively. A first order inertial response is added to both steering control and acceleration/deceleration control to formulate the system lag, where $\lambda_1>0$ and $\lambda_2>0$ are response coefficients. 

\begin{figure}
    \centering
    \includegraphics[width=8cm]{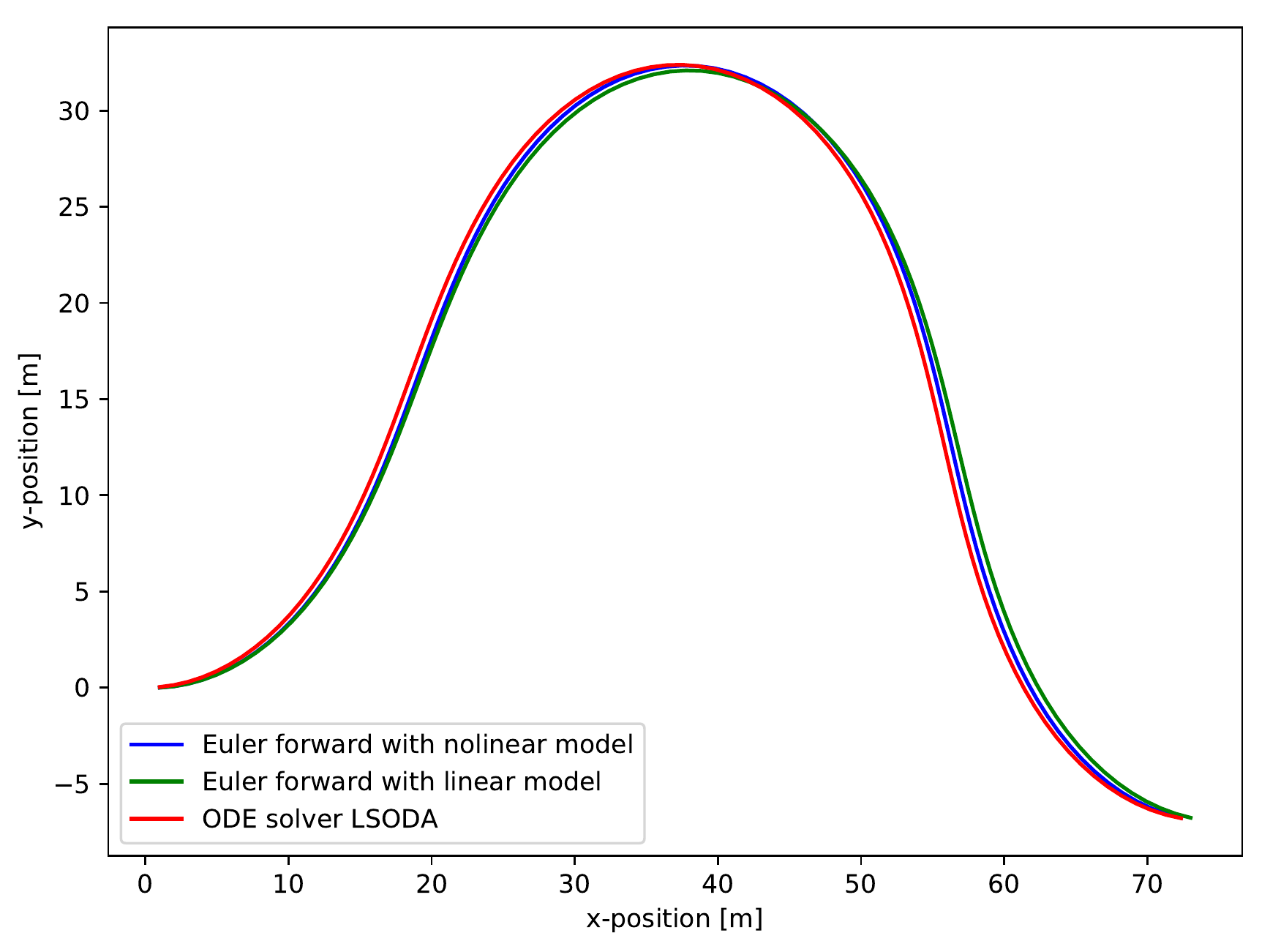}
    \caption{Verification of the linearized model. Initial condition is set as $[0,0,0,20,0]^T$ with $\alpha_{\mathrm{in}}$ and $\delta_{\mathrm{in}}$ sampled from sinusoid $2\cos(\frac{t}{5}\pi)$ and $0.2\cos(\frac{2t}{5}\pi)$ respectively, $0\leq t\leq 5$, $\Delta t=0.1s$, $\beta=0.5$. The red line shows the baseline generated using the ODE solver LSODA. The blue line and the green line denote the Euler forward approximation results of (\ref{eqn:vedyna}) and (\ref{eqn:linear}), respectively.}
    \label{fig:model_compare}
\end{figure}

Direct implementation of the nonlinear model (\ref{eqn:vedyna}) in the optimization (\ref{eqn:traj-opt}) results in a non-convex problem making real-time trajectory optimization intractable. 
It can be easily verified that even the one-step quadratic cost $x^TQx+\uff^TR\uff$ with (\ref{eqn:vedyna}) is non-convex in $\uff$ with both $Q$ and $R$ positive definite. 
In order to obtain a convex cost in $\Uff$, an LTV model is used to approximate (\ref{eqn:vedyna}). 
The linearized discrete-time LTV equation of (\ref{eqn:vedyna}) with $x\triangleq [s,y,\theta, \delta, v, \alpha]^T$ is given as follows.
\begin{equation}
\label{eqn:linear}
x(t+1)=A(t)x(t)+B(t)u(t)
\end{equation}
with $A(t)=$
\[
\begin{bmatrix}
1 & 0 & 0 & 0 & \cos{\theta}\Delta t & 0\\
0 & 1 & \beta v\Delta t & 0 & (1-\beta)\sin{\theta}\Delta t & 0\\
0 & 0 & 1 & \frac{v\Delta t}{L} & 0 & 0 \\
0 & 0 & 0 & 1 - \lambda_1 \Delta t & 0 & 0\\
0 & 0 & 0 & 0 & 1 & \Delta t\\
0 & 0 & 0 & 0 & 0 & 1-\lambda_2\Delta t
\end{bmatrix}
\]
\[
B(t)=B=\begin{bmatrix}
0 & 0 & 0 & \lambda_1\Delta t & 0 & 0\\
0 & 0 & 0 & 0 & 0 & \lambda_2\Delta t
\end{bmatrix}^T.
\]
where $\beta\in[0,1]$ is a weighting coefficient. The input matrix $B$ is time-invariant. Values of states in $A(t)$ are assigned using the corresponding states in $\hat{\tau}(t)$. Fig. \ref{fig:model_compare} shows numerical forward simulation of the LTV model (\ref{eqn:linear}) compared to (\ref{eqn:vedyna}).

\subsection{Convexity of Quadratic Functions}
\label{subsec:quad-prog}
In the interest of formulating (\ref{eqn:traj-opt}) as a SCOP, we focus on the quadratic programming formulations with strictly convex cost functions admitting unique solutions. 
Uniqueness of the optimal solution ensures that the optimized nominal trajectory is reproducible and solver-invariant\footnote{within the resolution tolerance set by the numerical solver.}. 
%
Quadratic programs are well understood and various numerical solvers are available~\cite{boyd2004convex}. 
While the focus is on quadratic programs, the synthesis approach proposed in this paper is applicable to SCOPs with cost functions of different categories. 
For instance, a SCOP with self-concordant barrier functions available on the constraints can be solved by a generic interior point method.
The following is a brief review of some useful facts and also introduces some of the notation that will be used in subsequent sections.
Throughout this work, let $\mathcal S_+^n$ (respectively, $\mathcal S_{++}^n$) be the set of positive semi-definite (respectively, positive definite) symmetric matrices in $\mathbb R^{n\times n}$. 
%
%
\begin{definition}[strict convexity]
\label{def:strict-convex}
Let $\mathcal X$ be a convex set in $\mathbb R^n$, the objective function $J:\mathcal X \rightarrow \mathbb R$ is called strictly convex if
$\forall x_1, x_2\in\mathcal{X}$, $\forall \lambda\in(0, 1)$, $J(\lambda x_1 + (1-\lambda) x_2) < \lambda J(x_1) + (1-\lambda) J(x_2)$.
\end{definition}

\begin{definition}[quadratic program]
A quadratic program over the decision variable $x\in\mathbb{R}^n$ is of the following form:
\begin{subequations}
\label{eqn:qp}
\begin{align}
    \text{minimize } & \frac{1}{2}x^THx + F^Tx + Y\\
    \text{subject to } & G_Ix\leq h\\
    & G_Ex = b
\end{align}
\end{subequations}
where $H\in\mathcal{S}_+^n$ is the Hessian matrix. $G_I\in\mathbb{R}^{m\times n}$, $G_E\in\mathbb{R}^{p\times n}$ are the element-wise inequality matrix and the element-wise equality matrix, respectively.
\end{definition}

If $H\in\mathcal{S}_{++}^{n}$ and the feasible set is convex, then the cost function is strictly convex and (\ref{eqn:qp}) is a strictly convex quadratic program so that the optimal solution $x^*$ exists and is unique. 
The notation $||x||_{R}^2$ indicates the quadratic form of $x$ where $R\in\mathcal{S}_{+}^{n}$. 

\section{MAIN RESULTS}
\label{sec:main-result}
\subsection{Cost Function Construction}
The dynamics of the LTV system over an $N$-step planning horizon is written compactly as
\begin{align}
\label{eqn:trajectory}
\mathring{\tau}(t) = \bm{A}(t)x(t) + \bm{B}(t)\Uff(t)
\end{align}
where 
\[
\bm{A}(t) = \begin{bmatrix}
I \\
A(t) \\
\vdots \\
\prod_{i=0}^{N-1}A(t+i)
\end{bmatrix},
\]
\[
\bf{B}(t) = \begin{bmatrix}
0 &  &  & \\
B &  &  & \\
A(t)B & \ddots &  & \\
\vdots &  & \ddots & \\
(\prod_{i=0}^{N-2}A(t+i))B & \cdots & \cdots & B
\end{bmatrix}.
\]

The idea of trajectory optimization is to construct a cost function to meet objectives leveraged by the motion planner and the feedback control system (refer to the connections in Fig. \ref{fig:sys-diagram}). For instance, the motion planner could have riding comfort as an objective and the feedback controller prefers slow changes over the nominal trajectory. It is observed that derivatives of control input are required in the objectives of optimization in addition to control input itself. Therefore, we focus on constructing the cost function of (\ref{eqn:qp}) that extends a generic MPC cost function with penalties on differentiation of control input in trajectory optimization. The difference matrix $E\in\mathbb{R}^{2N\times 2N}$ over the feedforward sequence $\Uff$ is as follows
\[
E = 
\begin{bmatrix}
I & & &  \\
-I & I & & \\
& \ddots & \ddots & \\
& & -I & I
\end{bmatrix},
\]
then the change of the control sequence is represented as 
\begin{equation}
    \Delta^{i+1} \Uff = E\Delta^{i}\Uff + V_{i}, \quad i\geq 0
\end{equation}
where $\Delta^{i}\Uff$ is the $i$-th order difference of $\Uff$ with $\Delta^{0}\Uff = \Uff$. $V_{i}$ is the estimation of the initial state that is stored at the previous step. For example, $V_0 = [-\uff(t-1)^T, 0,\dots,0]^T$ at sample $t$. 

Here we focus on various differences of control input since the changes of the state are encoded in the equations for the system dynamics. 
Given $R_i\in\mathcal{S}_{++}^{2N}$, we have
\begin{subequations}
\label{eqn:ith-order-cost}
\begin{align}
    ||\Delta^{i}\Uff||_{R_{i}}^2 = & ||E^i\Uff + \sum_{j=0}^{i-1}E^{j}V_{i-j-1}||_{R_{i}}^2\\
    = & ||\Uff||_{E^{i}{}^TR_{i}E^{i}}^2 + 2\left(\sum_{j=0}^{i-1}E^{j}V_{i-j-1}\right)^TR_{i}E^{i}\Uff\\
    & + ||\sum_{j=0}^{i-1}E^{j}V_{i-j-1}||_{R_{i}}^2.
\end{align}
\end{subequations}

The cost function comprising system state and control input is constructed as follows
\begin{align}
\label{eqn:cost}
    J(\mathring{\tau},\hat{\tau},\Uff) = ||\mathring{\tau}-\hat\tau||_{Q}^2 + ||\Uff||_{R_0}^2 + \sum_{i=1}^{M}||\Delta^{i}\Uff||_{R_i}^2
\end{align}
where $M\geq 0$ is the difference order of interest. $Q\in\mathcal{S}_{++}^{6(N+1)}$ is the weight coefficient of trajectory tracking deviation. $R_i\in\mathcal{S}_{++}^{2N}, 0\leq i\leq M,$ is the weight coefficient of the $i$-th order control input. If $M$ is set to $0$, (\ref{eqn:cost}) is equivalent to the cost function of an MPC scheme for tracking. In particular, we have the quadratic program of trajectory optimization as minimizing (\ref{eqn:cost}) subject to dynamic constraints (\ref{eqn:trajectory}), (\ref{eqn:traj-opt}c), and (\ref{eqn:traj-opt}d). Furthermore, we assume that the constraint sets $\mathcal{U}$ and $\mathcal{X}$ are polyhedra formulated by half planes generated by linear inequalities.

With the cost function designed in the form of (\ref{eqn:cost}), we have the following result on uniqueness.
\begin{proposition}
\label{prop:psd}
Given the trajectory optimization problem in the form of minimizing (\ref{eqn:cost}) subject to constraints (\ref{eqn:trajectory}), (\ref{eqn:traj-opt}c), and (\ref{eqn:traj-opt}d), if the polyhedral constraint set is non-empty, then the problem admits a unique optimal solution.
\end{proposition}
Justification of this observation can be found in Appendix \ref{app:lemma-psd}.

\subsection{Feedback and Feedforward Control}
When a new reference trajectory $\hat{\mathcal{T}}$ is available from the motion planner, the trajectory optimization takes the reference trajectory to set up the SCOP with (\ref{eqn:cost}), (\ref{eqn:linear}), and (\ref{eqn:traj-opt}c-d). The solution to the SCOP, $\Uff^{*}$, and the corresponding nominal trajectory together with a feedback controller form the feedback-feedforward control scheme. State update of the closed-loop system implementing the feedback-feedforward scheme is given as follows 
\begin{equation}
    x(t+1) = A(t)x(t) + B\uff^*(t) + A_{\mathrm{fb}}(t)\tilde{x}(t) + Bu_{\mathrm{fb}}(t)
\end{equation}
where $\uff^{*}\triangleq [\Uff^{*}(1), \Uff^{*}(2)]^T$, $\tilde{x} = x - \hat{x}$ is the state tracking error, $u_{\mathrm{fb}}$ is the corresponding feedback input. $A_{\mathrm{fb}}(t)=\frac{\partial f(x, u)}{\partial x}|_{x=\hat{x}(t)}$.

The state feedback controller follows the scheme of TVLQR. First, the system is augmented to include the integral of tracking error.
\begin{align}
    \tilde{z}(t+1) = \begin{bmatrix}
    A_{\mathrm{fb}}(t) & 0 \\
    C & I
    \end{bmatrix}\tilde{z}(t) + \begin{bmatrix}
    B \\
    0
    \end{bmatrix}u_{\mathrm{fb}}(t)
\end{align}
where $\tilde{z}=[\tilde{x}^T, \tilde{v}^T]^T$ is the augmented state with $\tilde{v}$ be the integrator state. The TVLQR problem is defined as follows
\begin{align}
    \min_{U_{\mathrm{fb}}} \sum_{k=0}^{N-1}\left(||\hat{z}(t+k)||_{\bar{Q}}^2 + ||u_{\mathrm{fb}}(t+k)||_{\bar{R}}^2\right) + ||\tilde{z}(t+N)||_{\bar{P}}^2
\end{align}
where $U_{\mathrm{fb}}(t)=[u_{\mathrm{fb}}^T(t),\dots,u_{\mathrm{fb}}^T(t+N-1)]^T$. $\bar{Q}\in\mathcal S_{+}^{12}$, $\bar{R}\in\mathcal{S}_{++}^{2}$, $\bar{P}\in\mathcal{S}_{++}^{12}$. Closed-loop stability and disturbance rejection properties of the feedback system can be found in \cite{rugh1996linear}. The trajectory optimization is called when a new trajectory from the motion planner is available. In general, the controller updates at a higher rate than the motion planner. Multiple samples in $\Uff$ are used in the feedback-feedforward scheme, which is different from solving a quadratic program per control step implemented in an MPC scheme. The TVLQR is responsible for disturbance rejection of the closed-loop system, which requires less computation resource.

\section{SIMULATION RESULTS AND DISCUSSION}
\label{sec:sim}
\subsection{Trajectory Optimization Results}
\label{subsec:traj-opt-result}
A sample trajectory is used to test the trajectory optimization. 
The trajectory consists of states $[s,y,\theta,v]^T$ over a 5-second horizon. 
The update interval of the trajectory optimization is set to $0.1s$ with the optimization horizon set to $5s$. 
The cost function is designed to minimize a weighted norm of the tracking error, the feedforward input, and the rate of change of feedforward input. 
In particular, $Q=\text{diag}(10, 10, 10, 0, 10, 0)$, $R_0=\text{diag}(0.1, 0.1)$, $R_1=\text{diag}(1,1)$. 
The constraint set $\mathcal{U}$ is set as $-4.0\leq\alpha_{\mathrm{in}}\leq 2.5$, and $-0.1\leq\delta_{\mathrm{in}}\leq 0.1$. Using the state model (\ref{eqn:trajectory}) with a 50 time-step horizon, the resulting quadratic program has 100 decision variables and 200 inequality constraints. 
The optimization problem is solved in MATLAB using $\mathbf{quadprog}$ on a Windows laptop with an Intel Core i5 CPU at 2.50GHz in 43.65 milliseconds averaged over 150 tests. 

\begin{figure}
    \centering
    \includegraphics[width=8cm]{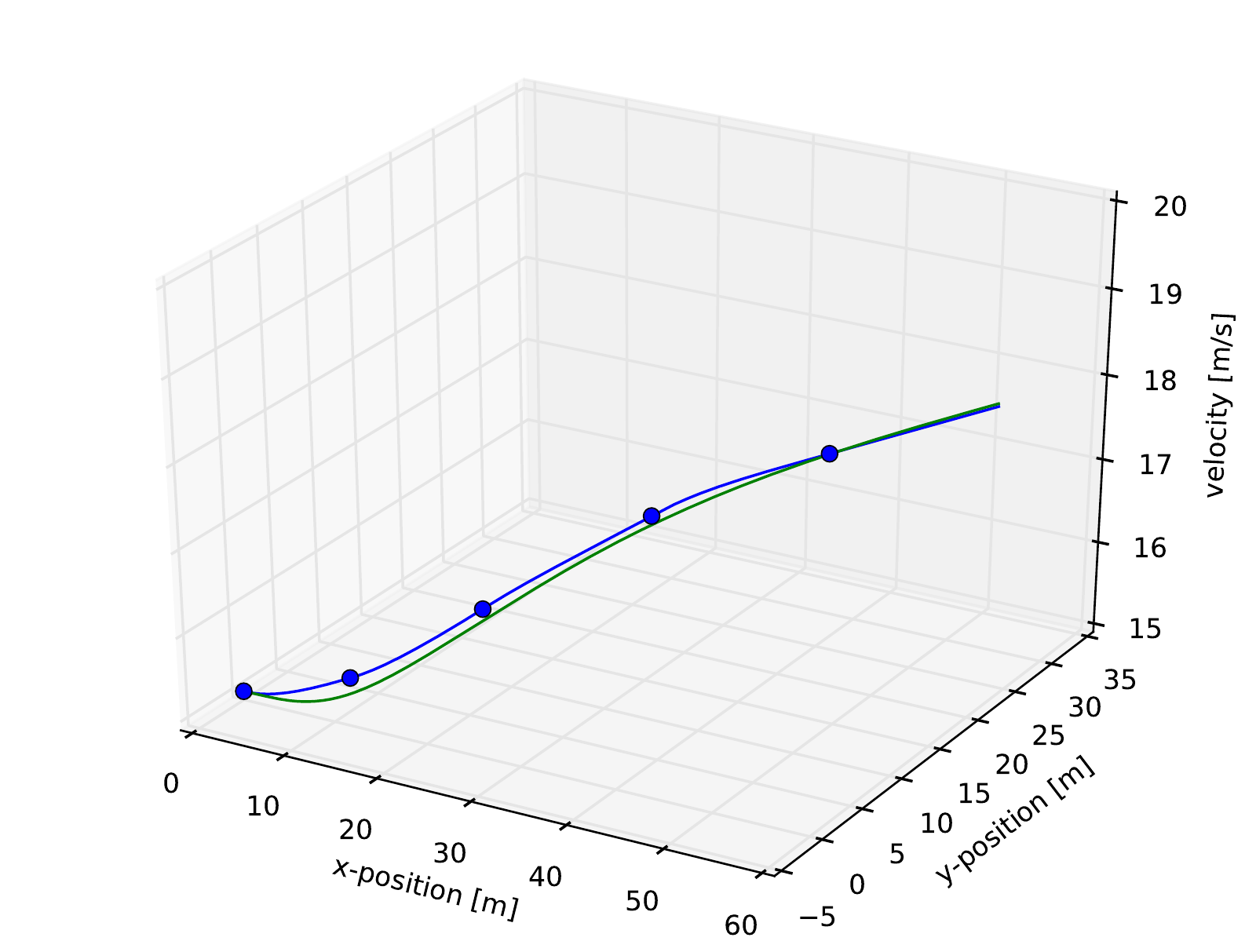}
    \caption{Optimization of a 5-second reference trajectory. The blue curve shows the coarse trajectory from the motion planner in the pose-velocity space, the green curve shows the optimized trajectory.}
    \label{fig:traj-compare}
\end{figure}

The reference trajectory from the planner is piecewise constant in steering angle and velocity consisting of 5 piecewise constant segments.
The reference velocity and steering angle over each sub-segment are constant and only change at the starting point of each sub-segment, denoted as blue dots in Fig. \ref{fig:traj-compare}. Fig. \ref{fig:traj-compare} shows the optimized trajectory in the $s-y-v$ space, depicted as the green curve. It is observed that the velocity over the optimized trajectory is adjusted in order to track the s-y position in the reference trajectory.

\begin{figure}
    \centering
    \includegraphics[width=8cm]{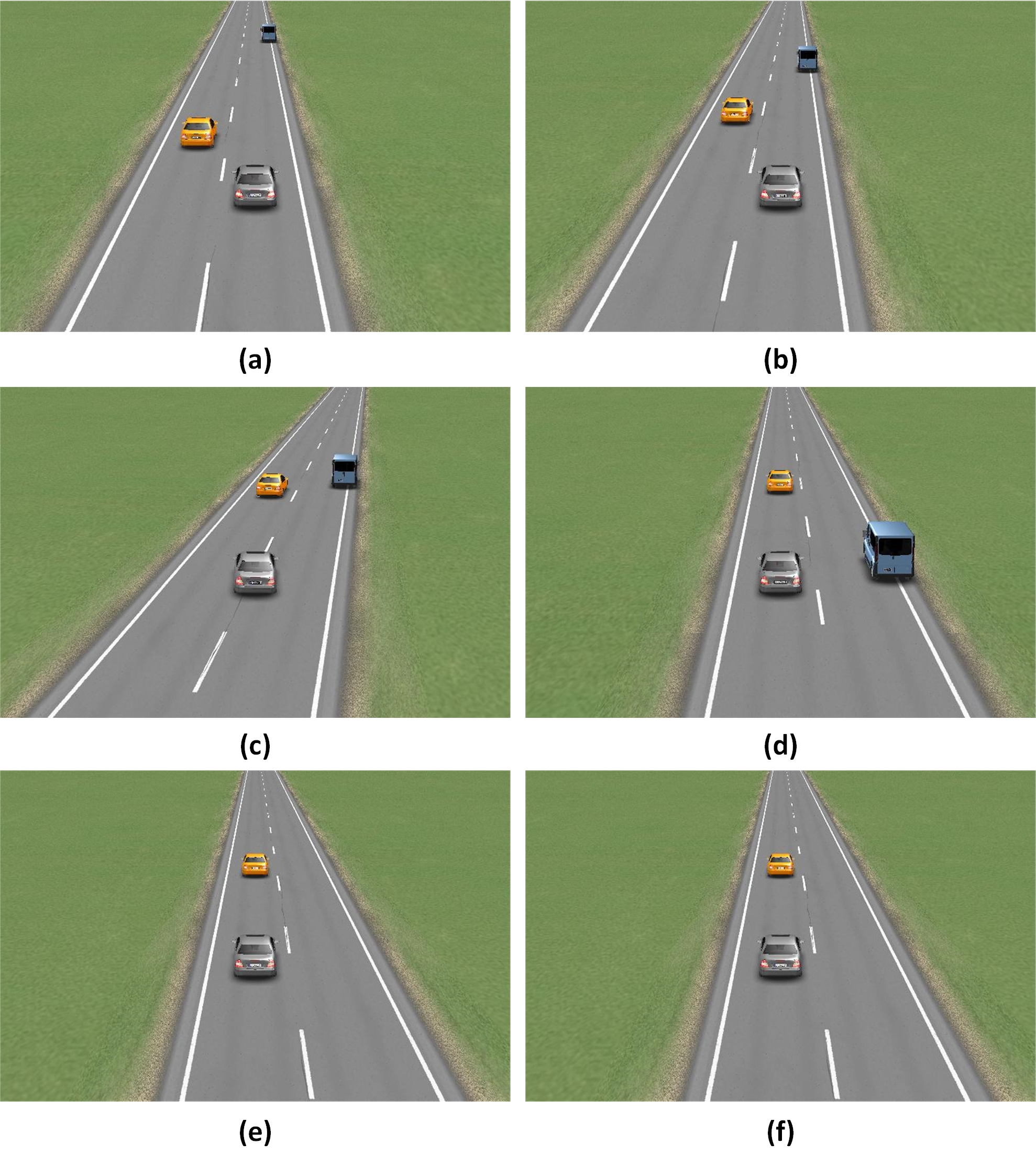}
    \caption{Screenshots of the simulation scenario. The grey sedan, the blue van, and the yellow sedan are the host vehicle, the pulled-over vehicle, and the cruising vehicle in the target lane, respectively.}
    \label{fig:carsim}
\end{figure}
\begin{figure}
    \centering
    \includegraphics[width=8cm]{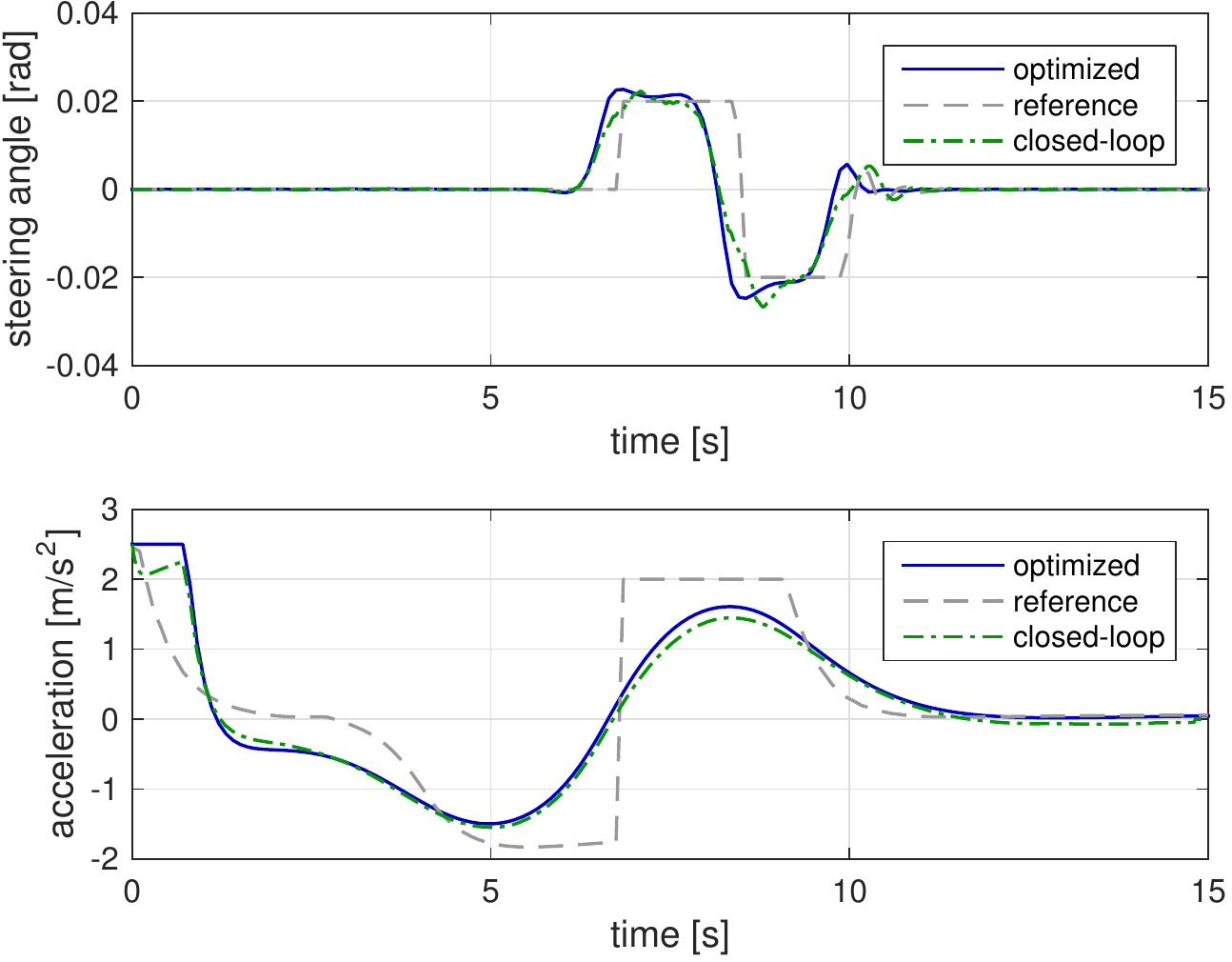}
    \caption{Control inputs during the evasion maneuver. Grey dash line: input derived from the reference trajectory; blue line: the feedforward input from the trajectory optimization; green dash-dot line: input of the closed-loop system.}
    \label{fig:feedforward}
\end{figure}

\subsection{Feedback-feedforward Performance}
In order to test the effectiveness of the proposed feedforward-feedback approach for motion trajectory tracking, a test scenario is designed considering an evasion maneuver. The host vehicle in the scenario needs to change lane to avoid a pulled-over vehicle while keeping safe inter-vehicle distance to the car in the target lane. The relative position of vehicles in the simulation is sequentially depicted in Fig. \ref{fig:carsim}.

The reference trajectory is obtained from the motion planner that minimizes deviations to a desired velocity profile calculated using time-headway and speed limits with forward kinematic simulation. The first 15-second of the planned trajectory is saved and is used as the reference trajectory. The dashed grey line in Fig. \ref{fig:feedforward} shows the estimated feedforward input derived from the reference trajectory using (\ref{eqn:linear}) without the first order lag. Steep changes of steering angle and acceleration are observed during $t=7$ and $t=10$ where the host vehicle needs to adjust longitudinal velocity while changing to the adjacent lane. Parameters of the trajectory optimization are set as the same to \ref{subsec:traj-opt-result}. Parameters of the TVLQR feedback controller is set as 
$\bar{Q}=\mathrm{diag}(10,5,10,1,10,1\mathrm{e}-4,1\mathrm{e}-4,0,0,0,0)$, $\bar{R}=\mathrm{diag}(1,5)$. The sampling interval of the TVLQR is set to $0.02s$ with the horizon set equal to the trajectory optimization horizon. The feedback-feedforward scheme is then tested on a nonlinear vehicle model that takes into account acceleration saturation of the powertrain, aero dynamics, and noises on steering and acceleration measurements.

\begin{figure}
    \centering
    \includegraphics[width=8cm]{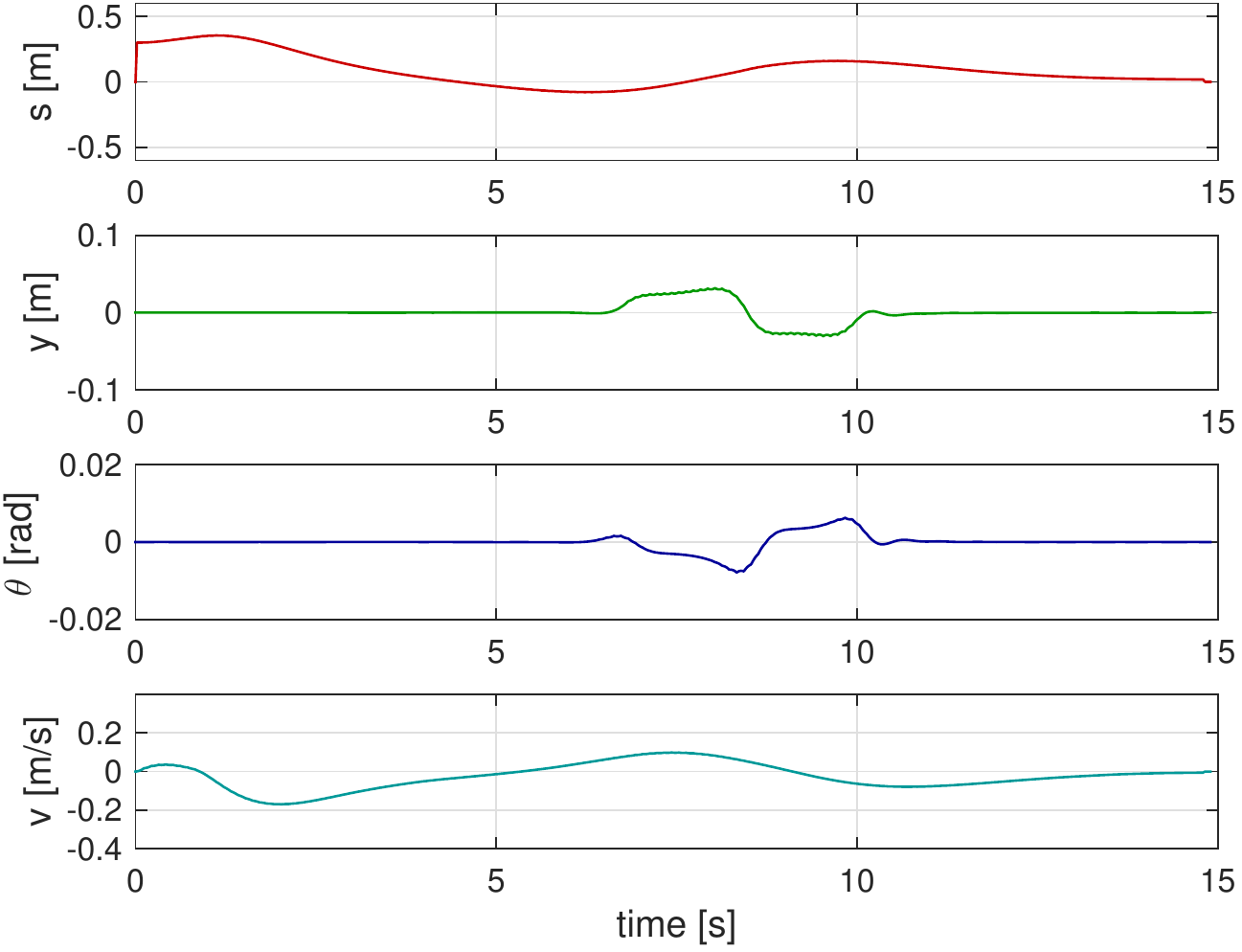}
    \caption{Trajectory tracking error of the feedback-feedforward scheme over the evasion maneuver.}
    \label{fig:track-err}
\end{figure}

Fig. \ref{fig:traj-compare} shows the tracking result of the feedback-feedforward scheme in $s-y$ plane. The corresponding input generated for the closed-loop system is shown in Fig. \ref{fig:feedforward} together with the feedforward term from the trajectory optimization. The feedforward term has been smoothed out during the evasion maneuver as the trajectory optimization takes into account i) change rates of the feedforward term, and ii) a vehicle model sharing the same state space as the feedback controller. Fig. \ref{fig:track-err} shows the tracking error over longitudinal position, lateral position, vehicle heading, and longitudinal velocity.

\begin{figure}
    \centering
    \includegraphics[width=8cm]{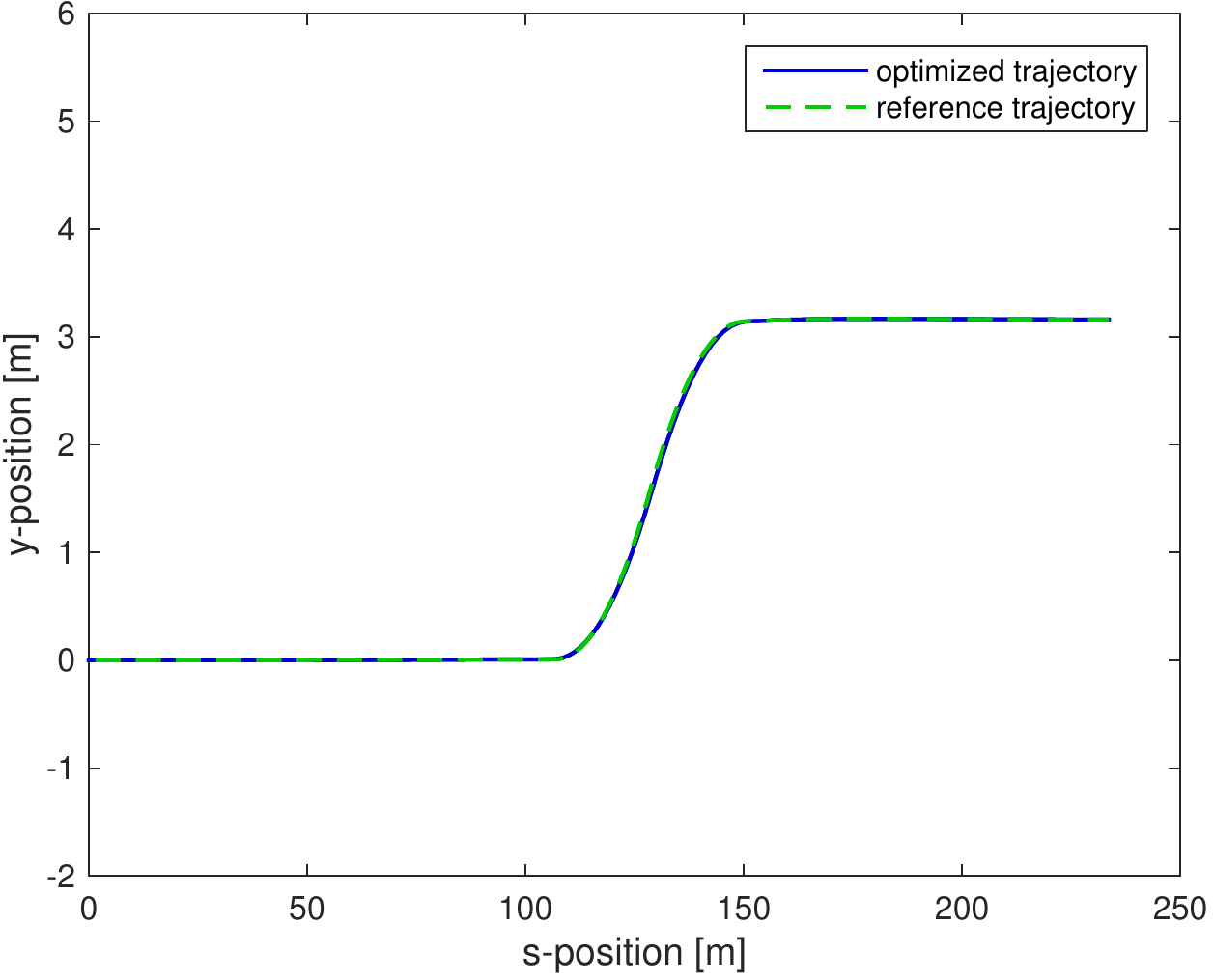}
    \caption{The tracking profile of the feedback-feedforward scheme in $s-y$ plane.}
    \label{fig:traj-group}
\end{figure}

\section{CONCLUSIONS}
\label{sec:conclude}
This paper proposes a trajectory tracking control approach for autonomous vehicles based on a model predictive trajectory optimization to generate a feedforward control and a time varying linear quadratic regulator for feedback.
%
%
Optimization of a reference trajectory is formulated as a strictly convex quadratic program by leveraging a linearization about the reference trajectory, polyhedral constraints, and a family of strictly convex quadratic cost functions. 
Additionally, the quadratic cost function is developed taking into account the rate of change of the feedforward input. 
A feedback-feedforward control scheme is proposed to actuate the vehicle by combining the optimized feedforward input and the feedback input generated by a TVLQR. 
The trajectory optimization and tracking scheme has been tested in simulation with an evasive maneuver. 
The proposed approach shows satisfactory tracking results in realistic driving scenarios, and with computation times well within the sample rate of the controller. 

\begin{appendices}
\section{Proof of Proposition \ref{prop:psd}}
\label{app:lemma-psd}
$R_i$ is a positive-definite symmetric matrix indicates that $R_i$ can be decomposed as $R_i=L_i^TL_i$. For any vector $x\in \mathbb R^{2N}$, $x^TE^i{}^TR_iE^ix = x^TE^i{}^TL_i^TL_iE^ix = L_iE^ix \cdot L_iE^ix \geq 0$. Therefore, $E^i{}^TR_iE^i \in \mathcal S_{+}^{n}$. Substituting $\mathring{\tau}(t)$ and $\Delta^i\Uff$ in (\ref{eqn:cost}) with (\ref{eqn:trajectory}) and (\ref{eqn:ith-order-cost}) respectively turns the cost function (\ref{eqn:cost}) into the following form.
\begin{align*}
    J &=||\bm{A}(t)x(t) + \bm{B}(t)\Uff(t) -\hat{\tau}||_Q^2 + ||\Uff||_{R_0}^2 + \prod_{i=1}^{M}||(\ref{eqn:ith-order-cost})||_{R_i}^2\\
    & =\Uff^TH\Uff + 2F^T\Uff + Y
\end{align*}
where $H = R_0 + \sum_{i=1}^{M}E^i{}^TR_{i}E^i + \bm{B}(t)^TQ\bm{B}(t)$, $F^T =\sum_{i=1}^{M}(\sum_{j=0}^{i-1}E^{j}V_{i-j-1})^TR_{i}E^{i} + (\bm{A}(t)x(t)-\hat{\tau})^TQ\bm{B}(t)$.

$E^i{}^TR_{i}E^i$, $\bm{B}(t)^TQ\bm{B}(t)$ are positive semidefinite and $R_0$ is positive definite, which indicates that $H$ is positive definite in $\Uff$. 
For any $U_1,U_2\in\mathbb{R}^{2N}, U_1\neq U_2$ and $\lambda \in (0, 1)$,
\begin{align*}
    & \lambda||U_1||_{H}^2 + (1-\lambda)||U_2||_{H}^2 - ||\lambda U_1 + (1-\lambda) U_2||_{H}^2 \\
  = & \lambda(1-\lambda)||U_1-U_2||_{H}^2
\end{align*}
where $\lambda(1-\lambda)>0$. Thus, $J$ is strictly convex in $\Uff$. 
Moreover, the non-empty polyhedral constraint set is a convex set. Therefore, the feasible solution in Proposition \ref{prop:psd} is a unique global optimum.

\end{appendices}

%




\bibliographystyle{IEEEtran}
\bibliography{IEEEabrv,RefListPeng}

%
%
%
%

\end{document}